\def\year{2020}\relax
\documentclass[letterpaper]{article} 
\usepackage{aaai20}  
\usepackage{times}  
\usepackage{helvet} 
\usepackage{courier}  
\usepackage[hyphens]{url}  
\usepackage{epsfig}
\usepackage{graphicx} 
\usepackage{amsmath}
\usepackage{amssymb}
\usepackage{amsthm}
\usepackage{booktabs}
\usepackage{multirow}
\usepackage{graphicx}  
\usepackage{subcaption}
\usepackage{array}
\newcommand{\PreserveBackslash}[1]{\let\temp=\\#1\let\\=\temp}
\newcolumntype{C}[1]{>{\PreserveBackslash\centering}p{#1}}
\newcolumntype{R}[1]{>{\PreserveBackslash\raggedleft}p{#1}}
\newcolumntype{L}[1]{>{\PreserveBackslash\raggedright}p{#1}}
\usepackage{pdfpages}
\title{Adversarial-Learned Loss for Domain Adaptation}
\author{
Minghao Chen, Shuai Zhao, Haifeng Liu, Deng Cai\thanks{Corresponding author}\\ 
State Key Lab of CAD\&CG, College of Computer Science, Zhejiang University, Hangzhou, China\\
Fabu Inc., Hangzhou, China\\
Alibaba-Zhejiang University Joint Institute of Frontier Technologies, Hangzhou, China\\
\{minghaochen01, zhaoshuaimcc\}@gmail.com, \{haifengliu, dcai\}@zju.edu.cn 
}
 
\begin{document}

\maketitle

\begin{abstract}
	Recently, remarkable progress has been made in learning transferable representation across domains. Previous works in domain adaptation are majorly based on two techniques: domain-adversarial learning and self-training. However, domain-adversarial learning only aligns feature distributions between domains but does not consider whether the target features are discriminative. On the other hand, self-training utilizes the model predictions to enhance the discrimination of target features, but it is unable to explicitly align domain distributions. In order to combine the strengths of these two methods, we propose a novel method called Adversarial-Learned Loss for Domain Adaptation (ALDA). We first analyze the pseudo-label method, a typical self-training method. Nevertheless, there is a gap between pseudo-labels and the ground truth, which can cause incorrect training. Thus we introduce the confusion matrix, which is learned through an adversarial manner in ALDA, to reduce the gap and align the feature distributions. Finally, a new loss function is auto-constructed from the learned confusion matrix, which serves as the loss for unlabeled target samples. Our ALDA outperforms state-of-the-art approaches in four standard domain adaptation datasets. Our code is available at \url{https://github.com/ZJULearning/ALDA}.
\end{abstract}

\begin{figure}[t]
	\begin{center}
		\includegraphics[width=1.0\linewidth]{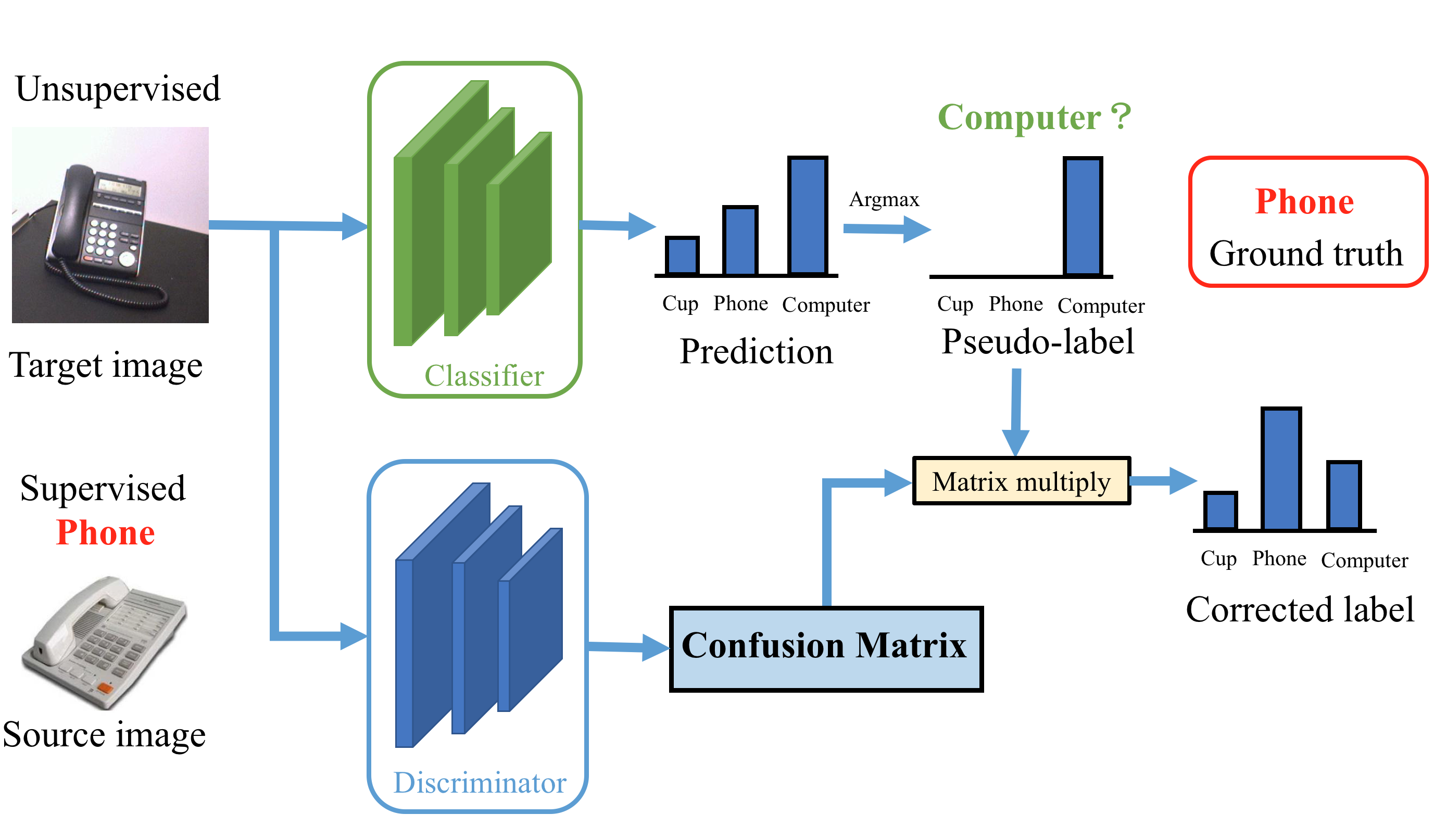}
	\end{center}
	\caption{The illustration of proposed adversarial-learned loss (ALDA). There is a gap between pseudo-label predicted by the model and the ground truth which is unavailable on the target domain. We employ a discriminator network to produce a confusion matrix to correct the pseudo-label, which then serves as the training label for the target sample. 
	}
	\label{fig_003}
\end{figure}

\section{Introduction} \label{Section_1}
In recent years, deep learning has made impressive progress in the classification task. The success of deep neural networks is based on the large scale datasets with a tremendous amount of labeled samples~\cite{ImageNet}. However, in many practical situations, a large number of labeled samples are inaccessible. The deep neural networks pre-trained on existing datasets cannot generalize well on the new data with different appearance characteristics. Essentially, the difference in data distribution between domains makes it difficult to transfer knowledge from the source to target domains. This transferring problem is known as \textit{domain shift}~\cite{domainshift}.

Unsupervised domain adaptation (UDA)
tackles the above \textit{domain shift} problem while transferring the model from a labeled source domain to an unlabeled target domain. The common idea of UDA is to make features extracted by neural networks similar between domains~\cite{DAN,DANN}. In particular, the domain-adversarial learning methods~\cite{DANN,ADDA} train a domain discriminator to distinguish whether the feature is from the source domain or target domain. To fool the discriminator, the feature generator has to output similar source and target feature distributions. 
However, it is challenging for this type of UDA methods to learn discriminative features on the target domain~\cite{MCD,LearningSemanticRepresentationsforUDA}. That is because they overlook whether the aligned target features can be discriminated by the classifier.

Recently, self-training based methods~\cite{self-ensembling,Self-training,Maxsquare} become another solution for UDA and achieve state-of-the-art performance on multiple tasks.
A typical way of self-training is to generate pseudo-labels corresponding to large prediction probability of target samples and train the model with these pseudo-labels. In this way, the features contributing to the target classification are enhanced. However, the alignment between the source and target feature distributions is implicit and has no theoretical guarantee. With unmatched target features, self-training based methods can lead to a drop of performance in the case of shallow networks~\cite{Self-training,MinimaxEntropy}. 

In conclusion, domain-adversarial learning is able to align the feature distributions with a theoretical guarantee, while self-training can learn discriminative target features. It is ideal to have a method to combine the advantages of these two types of methods. To achieve this
goal, we first analyze the loss function of self-training with pseudo-labels~\cite{Self-training} on the unlabeled target domain. Previous works in learning from noisy labels~\cite{LearningNL,Nosiylabel} proposed accounting for noisy labels with a confusion matrix. 
Following their analyzing approach, we reveal that the loss function using pseudo-labels~\cite{Self-training} differs from the loss function learned with the ground truth by a confusion matrix. Concretely, the commonly used
cross entropy loss
becomes:

\begin{align*}
& \mathcal{L}_{T}(x) =  \sum_{k=1}^K -p(y=k|x)\log p(\hat y=k|x)\\
= &\sum_{k=1}^K \sum_{l=1}^K -p(y=k|\hat y=l, x) p(\hat y=l|x)\log p(\hat y=k|x),
\end{align*}
where $K$ represents the number of categories, $y$ is the ground truth label for the sample $x$, $\hat y$ is the model prediction, i.e., pseudo-labels, and $p(y=k|\hat y=l, x)$ is the $(k, l)$-th component of the confusion matrix.

If the confusion matrix can be estimated correctly,  we can minimize the noise in pseudo-labels and boost the training of target samples. In this paper, we propose a novel method called Adversarial-learned Loss for Domain Adaptation (ALDA). As illustrated in Fig.~\ref{fig_003}, we generate the confusion matrix with a discriminator network. After multiplying with the confusion matrix, the pseudo-label vector turns into a corrected label vector, which serves as the training label on the target domain. As there is no direct way to optimize the confusion matrix, we learn it with \textit{noise-correcting domain discrimination}. Specifically, the domain discriminator has to produce different corrected labels for different domains, while the feature generator aims to confuse the domain discriminator. The adversarial process finally leads to a proper confusion matrix on the target domain.


The main contributions of this paper are as follows:
\begin{itemize}
	\item We analyze the noise in pseudo-labels with the confusion matrix, and propose our Adversarial-learned Loss for Domain Adaptation (ALDA) method, which uses adversarial learning to estimate the confusion matrix.
	
	\item We theoretically prove that ALDA can align the feature distributions between domains and correct the target prediction of the classifier. In this way, ALDA takes the strengths of domain-adversarial learning and self-training based methods. 
	
	\item ALDA can outperform state-of-the-art methods on four standard unsupervised domain adaptation datasets.
\end{itemize}

\section{Related Work}
\textbf{Unsupervised Domain Adaptation.}
With the success of deep learning, unsupervised domain adaptation (UDA)~\cite{DDC,DAN,JAN,DANN} has been embedded into deep neural networks to transfer the knowledge between the labeled source domain and unlabeled target domain. It has been revealed that the accuracy of the classifier on the target domain is bounded by the accuracy of the source and the domain discrepancy~\cite{Ben-David}. Therefore, the major line of the current UDA study is to align the distributions between the source and target domains. The distribution divergence between domains can be measured by Maximum Mean Discrepancy (MMD)~\cite{DDC,DAN} or second-order statistics~\cite{Deepcoral}.

\textbf{Domain-adversarial Methods.}
The domain-adversarial learning-based methods~\cite{DANN,ADDA} utilize a domain discriminator to represent the domain discrepancy. These methods play a minimax game: the discriminator is trained to distinguish the feature come from the source or target sample while the feature generator has to confuse the discriminator. 
However, due to practical issues, e.g., mode collapse~\cite{ModeRGA}, domain-adversarial learning cannot match the multi-modal distributions. Recently, together with the prediction of classifier~\cite{CADA,ConditionalGANDA}, the discriminator can match the distributions of each category, which significantly enhances the final classification results.

\textbf{Self-training Methods.} Semi-supervised learning~\cite{Semi-supervisedPseudo-Label,Semi-supervisedLearningbyEntropyMinimization,Meanteacher} is a similar task with domain adaptation, which also deals with labeled and unlabeled samples. With the data ``manifold'' assumption, some methods train the model based on the prediction of itself to smooth the decision boundary around the data. In particular, ~\cite{Semi-supervisedLearningbyEntropyMinimization} minimizes the prediction entropy as a regularizer for unlabeled samples. Pseudo-label method~\cite{Semi-supervisedPseudo-Label} selects high-confidence predictions as training target for unlabeled samples. Mean Teacher method~\cite{Meanteacher} sets the exponential moving average of the model as the teacher model and lets the prediction of the teacher model guide the original model.

Recently, many works apply the above self-training based methods to unsupervised domain adaptation~\cite{Self-training,Maxsquare,self-ensembling}. These UDA methods implicitly encourage the class-wise feature alignment between domains and achieve surprisingly good results on multiple UDA tasks.

\begin{figure*}[t]
	\begin{center}
		\includegraphics[width=0.9\linewidth]{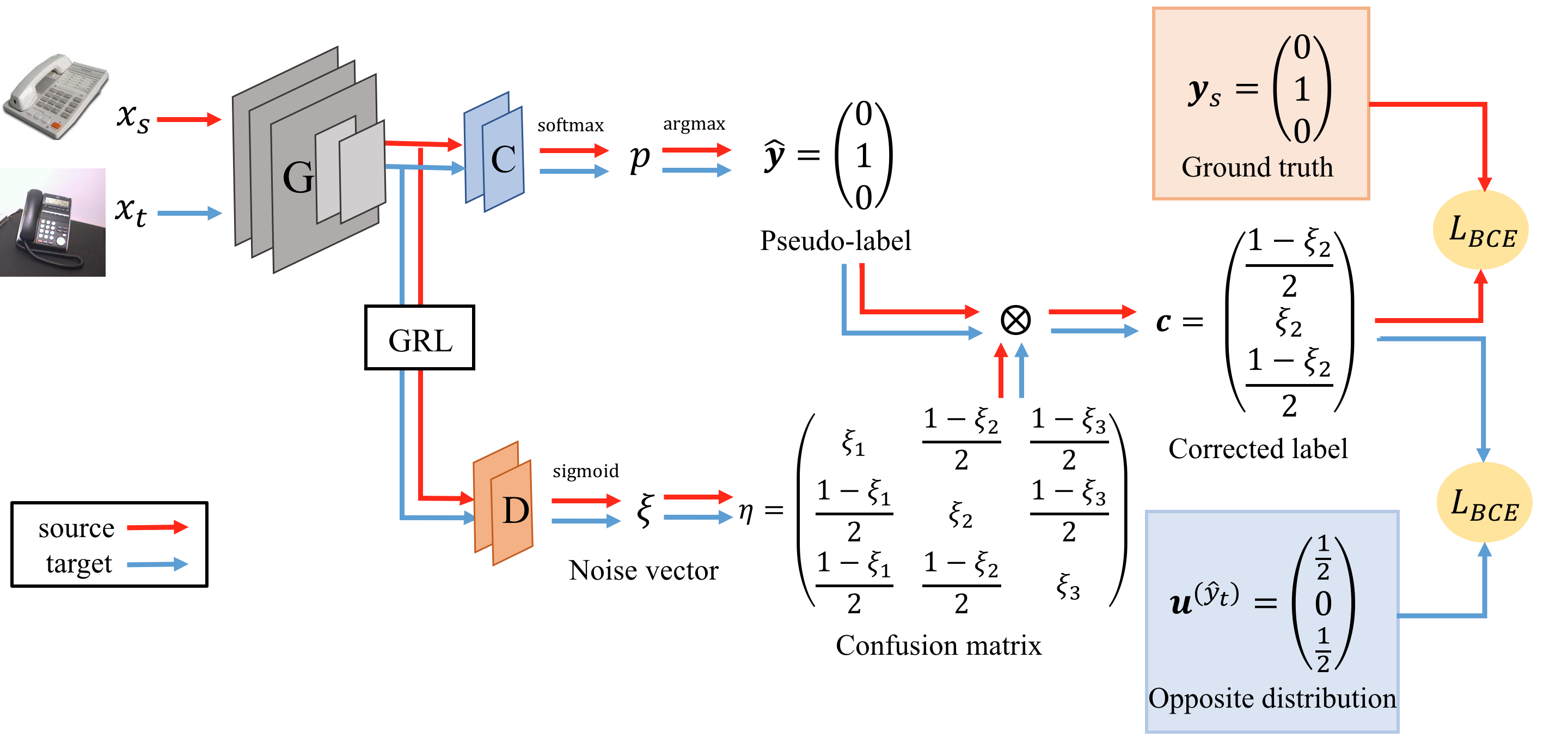}
	\end{center}
	\caption{The illustration of noise-correcting domain discrimination ($K=3$). The confusion matrix $\eta$ is class-wise uniform with the vector $\xi$ generated by the discriminator $D$. The corrected pseudo-label $\mathbf{c}$ is generated by multiplying the confusion matrix $\eta$ and the pseudo-label vector $\mathbf{\hat{y}}$. For the source sample, the target of $\mathbf{c}$ is the ground truth $\mathbf{y_s}$, and the target is the opposite distribution for the target sample. The generator $G$ is designed to confuse the above targets. Therefore, we add a gradient reverse layer (GRL)~\cite{DANN} to achieve the minimax optimization.}
	\label{fig_001}
\end{figure*}

\section{Methods}

\subsection{Preliminaries}
For unsupervised domain adaptation, we have a labeled source domain $\mathcal{D}_S=\{(x^i_s, y^i_s)\}_{i=1}^{n_s}$ and a unlabeled target domain $\mathcal{D}_T=\{x^j_t \}_{j=1}^{n_t}$. We train a generator network $G$ to extract the high-level feature from the data $x_s$ or $x_t$, and a classifier network $C$ to finish the $K$-class classification task on the feature space. The classifier $C$ outputs probability vectors $\mathbf{p}_s, \mathbf{p}_t\in\mathbb { R } ^ { K }$, indicating the prediction probability of $x_s, x_t$ respectively. 

In this paper, we consider providing a proper loss function on the target domain. Theoretically, the ideal loss function is the loss with the ground truth $y_t$: 
\begin{align}
\mathcal{L}_{T}(x_t, \mathcal{L})=\sum_{k=1}^K p(y_t=k|x_t) \mathcal{L}(\mathbf{p}_t, k),
\end{align}
where $\mathcal{L}$ is a basic loss function, e.g., cross entropy (CE), mean absolute error (MAE). 

However, the target ground truth $y_t$ is unavailable in the UDA setting. Pseudo-label method~\cite{Semi-supervisedPseudo-Label,Self-training} substitutes $y_t$ with the model prediction: $\hat y_t=\mathrm{argmax}_k\,{\mathbf{p}^k_t}, \text{if}\,\max_k\,{\mathbf{p}^k_t}>\delta$, where $\delta$ is a threshold. As mentioned in the introduction, we analyze the difference between the ideal loss and the loss with pseudo-labels:

\begin{align}
& \mathcal{L}_{T}(x_t, \mathcal{L}) =  \sum_{k=1}^K  p(y_t=k|x_t) \mathcal{L}(\mathbf{p}_t, k)\\
= &\sum_{k=1}^K \sum_{l=1}^K p(y_t=k|\hat y_t=l, x_t) p(\hat y_t=l|x_t) \mathcal{L}(\mathbf{p}_t, k) \\
= &\sum_{k=1}^K \sum_{l=1}^K \eta_{kl}^{(x_t)} p(\hat y_t=l|x_t) \mathcal{L}(\mathbf{p}_t, k),
\end{align}
where $\eta^{(x_t)}$ is the confusion matrix. The confusion matrix is unknown on the unlabeled target domain.
For brevity, we define $\mathbf{c}^{(x_t)}_k= \sum_l \eta_{kl}^{(x_t)} p(\hat y_t=l|x_t)$ and name $\mathbf{c}^{(x_t)}$ as the corrected label vector.

In previous works studying noisy labels~\cite{Nosiylabel}, it is commonly assumed that the confusion matrix is conditionally independent of inputs $x_t$ and \textit{uniform} with noise rate $\alpha$. The unhinged loss has been proved to be robust to the \textit{uniform} noise~\cite{Unhinge,RobustLF}, 

\begin{align}
\mathcal{L}_{unh}(\mathbf{p}, k) = 1-\mathbf{p}_k. \label{equ_007}
\end{align}

However, these assumptions
cannot hold in the case of pseudo-labels, which makes the problem more intractable.

\subsection{Adversarial-Learned Loss}

The general idea of our method is that if we can adequately estimate the noise matrix $\eta_{kl}^{(x_t)}$, the noise in pseudo-labels will be corrected and we can approximately optimize the ideal loss function on the target domain. 

Firstly, to simplify the noisy label problem, we assume that the noise is class-wise uniform with vector $\xi^{(x_t)}$. 

\textbf{Definition 1.} Noise is \textit{class-wise uniform} with vector $\xi^{(x_t)}\in\mathbb { R } ^ { K }$, if $\eta_{kl}^{(x_t)}=\xi_k^{(x_t)}$ for $k=l$, and $\eta_{kl}^{(x_t)}=\frac{1-\xi_l^{(x_t)}}{K-1}$ for $k\ne l$. 

In this work, we propose to use an extra neural network, called noise-correcting domain discriminator, to learn the vector $\xi^{(x_t)}$.

\subsection{Noise-correcting Domain Discrimination}

As shown in Fig.~\ref{fig_001}, the noise-correcting domain discriminator $D$ is a multi-layer neural network, which takes the deep feature $G(x)$ as the input and outputs a multi-class score vector $D(G(x))\in\mathbb { R } ^ { K }$. After a sigmoid layer, the discriminator produces the noise vector $\xi^{(x)} = \sigma(D(G(x)))$. Each component of $\xi^{(x)}$ denotes the probability that the pseudo label is the same as the correct label: $\xi^{(x)}_k=p(y=k|\hat y=k, x)$.

We adopt the idea of the domain-adversarial learning~\cite{DANN} that makes the discriminator and the generator play a minimax game. Instead of letting the discriminator perform a domain classification task, we let the discriminator generate different noise vectors for the source and target domains.
As illustrated in Fig.~\ref{fig_001}, for the source feature $G(x_s)$, the discriminator aims to minimize the discrepancy between the corrected label vector $\mathbf{c}^{(x_s)}$ and the ground truth $\mathbf{y}_s(=\mathrm{one\_hot}(y_s))$. The adversarial loss for the source data is:

\begin{align}
\mathcal{L}_{Adv}(x_s, y_s) &= \mathcal{L}_{BCE}(\mathbf{c}^{(x_s)}, \mathbf{y}_s)  \\
= \sum_k -{\mathbf{y}_s}_k\log \mathbf{c}^{(x_s)}_k &- (1-{\mathbf{y}_s}_k)\log(1- \mathbf{c}^{(x_s)}_k).
\end{align}

As for the target feature $G(x_t)$, the discriminator do the opposite way. The discriminator will correct pseudo-labels to the opposite distribution $\mathbf{u}^{(\hat y_t)}\in\mathbb { R } ^ { K }$, in which $\mathbf{u}^{(\hat y_t)}_k = 0$ for $k=\hat y_t$ and $\mathbf{u}^{(\hat y_t)}_k = \frac{1}{(K-1)}$ for $k\ne\hat y_t$. The adversarial loss for the target data is:

\begin{align}
\mathcal{L}_{Adv}(x_t) = \mathcal{L}_{BCE}(\mathbf{c}^{(x_t)}, \mathbf{u}^{(\hat y_t)}).
\end{align}

The total adversarial loss becomes:
\begin{align}
\mathcal{L}_{Adv}(x_s, y_s, x_t) = \mathcal{L}_{Adv}(x_s, y_s) + \mathcal{L}_{Adv}(x_t).
\end{align}

The discriminator $D$ needs to minimize the loss function to distinguish between the source and target feature. On the other hand, the generator $G$ has to fool the discriminator, by maximizing the above loss function. Compared to the common domain-adversarial learning, this adversarial loss takes the classifier prediction and the label information into consideration. In this way, our noise-correcting domain discriminator can achieve the class-wise feature alignment.

\subsection{Regularization Term} \label{section_reg}

As revealed in the works of generative adversarial networks (GANs)~\cite{LSGAN}, the training process of adversarial learning can be unstable. Following ~\cite{ACGAN}, we add a classification task on the source domain to the discriminator to make its training
more stable. Consequently, the discriminator not
only has to distinguish the source and target domains but also correctly classify the source samples.


To embed the classification task into training, we add a regularization term to the loss of the discriminator:
\begin{align}
\mathcal{L}_{Reg}(x_s, y_s) = \mathcal{L}_{CE}(\mathbf{p}_D^{(x_s)}, y_s), \label{equ_012}
\end{align}
where $\mathbf{p}_D^{(x_s)} = \mathrm{softmax}(D(G(x_s)))$ and $\mathcal{L}_{CE}$ is the cross entropy loss. Then the final loss function for the discriminator becomes:

\begin{align}
\min_D E_{(x_s, y_s), x_t}(\mathcal{L}_{Adv}(x_s, y_s, x_t)+\mathcal{L}_{Reg}(x_s, y_s)).
\end{align}

\subsection{Corrected Loss Function}

After the adversarial learning of the confusion matrix $\eta^{(x_t)}$, we can 
construct a proper loss function for the target samples. As the unhinged loss (Eq.~\ref{equ_007}) is robust to the uniform part of noise, we choose the unhinged loss $\mathcal{L}_{unh}$ as the basic loss function $\mathcal{L}$:

\begin{align}
\mathcal{L}_{T}(x_t, \mathcal{L}_{unh})&=\sum_{k,l} \eta_{kl}^{(x_t)} p(\hat y_t=l|x_t) \mathcal{L}_{unh}(\mathbf{p}_t, k) \\
&=\sum_{k} \mathbf{c}^{(x_t)}_k\mathcal{L}_{unh}(\mathbf{p}_t, k). \label{equ_015}
\end{align}

Together with the supervised loss on the source domain, the losses for the classifier and the generator become:

\begin{align}
\min_C E_{(x_s, y_s), x_t}(\mathcal{L}_{CE}(p_s, y_s)+\lambda\mathcal{L}_{T}(x_t, \mathcal{L}_{unh}))
\end{align}

\begin{align}
\min_G E_{(x_s, y_s), x_t}(&\mathcal{L}_{CE}(p_s, y_s)+\lambda\mathcal{L}_{T}(x_t, \mathcal{L}_{unh}) \nonumber\\
&-\lambda\mathcal{L}_{Adv}(x_s, y_s, x_t)),
\end{align}
where $\lambda\in[0,1]$ is a trade-off parameter.

\section{Theoretical Insight}

In the feature space $\mathcal{F}$ generated by the generator $G$, the source and target feature distributions are $\mathcal{P}_s=\{G(x_s)|x_s\in\mathcal{D}_s\}$ and $\mathcal{P}_t=\{G(x_t)|x_t\in\mathcal{D}_t\}$ respectively. If we assume that both distributions are continuous with densities $P_s$ and $P_t$, for a feature vector $f\in\mathcal{F}$, the probabilities that it belongs to source and target distributions are $P_s(f)$ and $P_t(f)$ respectively. 

\subsubsection{Theorem 1.} \textit{When the noise-correcting domain discrimination}
\begin{align}
\max_G\min_DE_{(x_s, y_s), x_t}\mathcal{L}_{Adv}(x_s, y_s, x_t)
\end{align}
\textit{achieves the optimal point $D^*$ and $G^*$, the feature distributions generated by $G^*$ are aligned: $\mathcal{P}_s=\mathcal{P}_t$.}

\textit{Proof.} The proof is given in the supplemental material.

As a result, the noise-correcting domain discrimination can align the feature distribution between the source and target domain. According to the theory of~\cite{Ben-David}, the expected error on the target samples can be bounded by the expected error on the source domain and feature discrepancy between domains. Therefore, the target expected error of our noise-correcting domain discrimination is theoretically bounded.

\begin{table*}[t]
	\centering
	\resizebox{0.75\textwidth}{21.0mm}{    
		\begin{tabular}{cccccccc}  
			\toprule
			Method & A $\to$ W & D $\to$ W &  W $\to$ D & A $\to$ D & D $\to$ A&  W $\to$ A & Avg    \\
			\midrule
			ResNet-50~\cite{ResNet} &  68.4$\pm$0.2 & 96.7$\pm$0.1  & 99.3$\pm$0.1  & 68.9$\pm$0.2 & 62.5$\pm$0.3 & 60.7$\pm$0.3 & 76.1    \\
			DANN~\cite{DANN}  &  82.0$\pm$0.4 & 96.9$\pm$0.2  & 99.1$\pm$0.1  & 79.7$\pm$0.4 & 68.2$\pm$0.4 & 67.4$\pm$0.5 & 82.2    \\
			ADDA~\cite{ADDA}   &  86.2$\pm$0.5 & 96.2$\pm$0.3 & 98.4$\pm$0.3 & 77.8$\pm$0.3 & 69.5$\pm$0.4 & 68.9$\pm$0.5 & 82.9 \\
			JAN~\cite{JAN}  & 85.4$\pm$0.3 & 97.4$\pm$0.2 & 99.8$\pm$0.2 & 84.7$\pm$0.3 & 68.6$\pm$0.3 & 70.0$\pm$0.4 & 84.3 \\
			MADA~\cite{MADA}  & 90.0$\pm$0.1 & 97.4$\pm$0.1 & 99.6$\pm$0.1 & 87.8$\pm$0.2 & 70.3$\pm$0.3 & 66.4$\pm$0.3 & 85.2 \\
			CBST~\cite{Self-training} & 87.8$\pm$0.8 & 98.5$\pm$0.1 & \textbf{100$\pm$0.0} & 86.5$\pm$1.0 & 71.2$\pm$0.4 & 70.9$\pm$0.7 & 85.8 \\
			CAN~\cite{CAN} & 92.5 & \textbf{98.8} & \textbf{100.0} & 90.1 & 72.1 & 69.9 & 87.2 \\
			CDAN+E~\cite{CADA} & 94.1$\pm$0.1 & 98.6$\pm$0.1 & \textbf{100.0$\pm$0.0} & 92.9$\pm$0.2 & 71.0$\pm$0.3 & 69.3$\pm$0.3 & 87.7 \\
			MCS~\cite{MCS} & - & - & - & - & - & - & 87.8 \\
			\midrule
			\textbf{ALDA} & \textbf{95.6$\pm$0.5} & 97.7$\pm$0.1 & \textbf{100.0$\pm$0.0} & \textbf{94.0$\pm$0.4} & \textbf{72.2$\pm$0.4} & \textbf{72.5$\pm$0.2} & \textbf{88.7} \\
			\bottomrule
	\end{tabular}}
	\caption{Accuracy (\%) of different unsupervised domain adaptation methods on Office-31 (ResNet-50)}
	\label{tab_002}
\end{table*}

\begin{table*}[t]
	\centering
	\resizebox{1.0\textwidth}{14.3mm}{    
		\begin{tabular}{clc}  
			\toprule
			Method & Ar $\to$ Cl Ar $\to$ Pr Ar $\to$ Rw Cl $\to$ Ar Cl $\to$ Pr Cl $\to$ Rw\,Pr $\to$ Ar Pr $\to$ Cl Pr $\to$ Rw Rw $\to$ Ar Rw $\to$ Cl\,Rw $\to$ Pr  & Avg    \\
			\midrule
			ResNet-50~\cite{ResNet} &\quad 34.9\qquad 50.0\qquad 58.0\qquad37.4\qquad 41.9\qquad 46.2\qquad 38.5\qquad 31.2\qquad 60.4       \,\,\,\qquad 53.9       \,\qquad 41.2      \,\qquad 59.9& 46.1    \\
			DANN~\cite{DANN}  &\quad 45.6\qquad 59.3\qquad 70.1\qquad 47.0\qquad 58.5\qquad 60.9\qquad 46.1\qquad 43.7\qquad 68.5      \,\,\,\qquad 63.2       \,\qquad 51.8      \,\qquad 76.8& 57.6    \\
			JAN~\cite{JAN}  &\quad 45.9\qquad 61.2\qquad 68.9\qquad 50.4\qquad 59.7\qquad 61.0\qquad 45.8\qquad 43.4\qquad 70.3      \,\,\,\qquad 63.9      \,\qquad 52.4      \,\qquad 76.8& 58.3 \\
			CDAN+E~\cite{CADA} &\quad 50.7\qquad \textbf{70.6}\qquad 76.0\qquad 57.6\qquad 70.0\qquad 70.0\qquad 57.4\qquad 50.9\qquad \textbf{77.3}     \,\,\,\qquad \textbf{70.9}      \,\qquad \textbf{56.7}      \,\qquad 81.6& 65.8 \\
			TAT~\cite{TAT}          &\quad 51.6\qquad 69.5\qquad 75.4\qquad 59.4\qquad 69.5\qquad 68.6\qquad \textbf{59.5}\qquad 50.5\qquad 76.8      \,\,\,\qquad \textbf{70.9}      \,\qquad 56.6     \,\qquad 81.6& 65.8 \\
			\midrule
			\textbf{ALDA} &\quad \textbf{53.7}\qquad 70.1\qquad \textbf{76.4}\qquad \textbf{60.2}\qquad \textbf{72.6}\qquad \textbf{71.5}\qquad 56.8\qquad \textbf{51.9}\qquad 77.1      \,\,\,\qquad 70.2      \,\qquad 56.3      \,\qquad 82.1& \textbf{66.6}
			\\
			\bottomrule
	\end{tabular}}
	\caption{Accuracy (\%) of different unsupervised domain adaptation methods on Office-Home (ResNet-50)}
	\label{tab_003}
\end{table*}

Furthermore, we can prove that by optimizing the corrected loss function, the noise in pseudo-labels is reduced.

\subsubsection{Theorem 2.}
\textit{When the optimal point $D^*$ and $G^*$ are achieved in Theorem 1, if there is a optimal labeling function $y^*(f_s)=y_s, \forall f_s \in \mathcal{P}_s$ in the feature space $\mathcal{F}$, then $\forall x_t \in \mathcal{P}_t$ and $f_t=G^*(x_t)$, we have:
	\begin{align*}
	\mathbf{c}^{(x_t)} =  \begin{cases} 
	\mathbf{h}^{y^*(f_t)}  & \hat y_t=y^*(f_t)  \\  
	\mathbf{u}^{(\hat y_t)}  & \text { otherwise }, \end{cases}
	\end{align*}
	where $\mathbf{c}^{(x_t)}=\mathbf{h}^{y^*(f_t)}$ denotes that $\mathbf{c}^{(x_t)}_k = \frac{1}{2}$ for $k=\hat y_t$ and $\mathbf{c}^{(x_t)}_k = \frac{1}{2K-2}$ otherwise. }

\textit{Proof.} The proof is given in the supplemental material.

As Theorem 2 shows, when we optimize the target loss $\mathcal{L}_{T}(x_t, \mathcal{L})=\sum_{k} \mathbf{c}^{(x_t)}_k\mathcal{L}(p_t, k)$, the loss of pseudo-labels $\mathcal{L}(p_t, \hat y_t)$ will be enhanced when $\hat y_t=y^*(x_t)$ ($\mathbf{c}^{(x_t)}_{\hat y_t}=\frac{1}{2}$) and suppressed otherwise ($\mathbf{c}^{(x_t)}_{\hat y_t}=0$). In this way, the training of classifier can be corrected by the discriminator on the target domain and will be more efficient than the original pseudo-label method.

\section{Experiments}
We evaluate the proposed adversarial-learned loss for domain adaptation (ALDA) with state-of-the-art approaches on four standard unsupervised domain adaptation datasets: digits, office-31, office-home, and VisDA-2017. 

\subsection{Datasets}
\textbf{Digits.} Following the evaluation protocol of~\cite{CADA}, we experiment on three adaptation scenarios: USPS to MNIST (\textbf{U} $\to$ \textbf{M}), MNIST to USPS (\textbf{M} $\to$ \textbf{U}), and SVHN to MNIST (\textbf{S} $\to$ \textbf{M}). MNIST~\cite{MINST} contains $60,000$ images of handwritten digits and USPS~\cite{USPS} contains $7,291$ images. Street View House Numbers (SVHN)~\cite{SVHN} consists of $73,257$ images with digits and numbers in natural scenes. We report the evaluation results on the test sets of MNIST and USPS.

\textbf{Office-31} ~\cite{office} is a commonly used dataset for unsupervised domain adaptation, which contains $4,652$ images and $13$ categories collected from three domains: \textit{Amazon} (\textbf{A}), \textit{Webcam} (\textbf{W}) and \textit{DSLR} (\textbf{D}). We evaluate all methods across six domain adaptation tasks: \textbf{A} $\to$ \textbf{W}, \textbf{D} $\to$ \textbf{W}, \textbf{W} $\to$ \textbf{D}, \textbf{A} $\to$ \textbf{D}, \textbf{D} $\to$ \textbf{A} and \textbf{W} $\to$ \textbf{A}.

\textbf{Office-Home}~\cite{officehome} is a more difficult domain adaptation dataset than office-31, including $15,500$ images from four different domains: Artistic images (\textbf{Ar}), Clip Art (\textbf{Cl}), Product images (\textbf{Pr}) and Real-World (\textbf{Rw}). For each domain, the dataset contains images of $65$ object categories that are common in office and home scenarios. We evaluate all methods in $12$ adaptation scenarios.

\textbf{VisDA-2017}~\cite{VisDA} is a large-scale dataset and challenge for unsupervised domain adaptation from simulation to real. The dataset contains $152,397$ synthetic images as the source domain and $55,388$ real-world images as the target domain. $12$ object categories are shared by these two domains. Following previous works~\cite{MCD,CADA}, we evaluate all methods on the validation set of VisDA.

\begin{table*}[t]
	\centering
	\resizebox{0.95\textwidth}{17.0mm}{    
		\begin{tabular}{ccccccccccccccc}  
			\toprule
			Method & Backbone & plane & bcycl & bus & car & house &  knife & mcycl & person & plant & sktbrd & train & truck & Avg    \\
			\midrule
			Sourceonly & \multirow{5}{*}{ResNet101} &  55.1 &    53.3 &    61.9 &    59.1 &    80.6 &    17.9 &    79.7 &    31.2 &    81.0 &    26.5 &    73.5 &    8.5 & 52.4 \\
			DANN~\cite{DANN} & &  81.9 & \textbf{77.7} & 82.8 & 44.3 & 81.2 & 29.5 & 65.1 & 28.6 & 51.9 & 54.6 & 82.8 & 7.8 & 57.4    \\
			MCD~\cite{MCD} &  & 87.0 & 60.9 & \textbf{83.7} & 64.0 & 88.9 & 79.6 & 84.7 & \textbf{76.9} & 88.6 & 40.3 & 83.0 & 25.8 & 71.9 \\
			CBST~\cite{Self-training}  & & 87.2 & 78.8 & 56.5 & 55.4 & 85.1 & 79.2 & 83.8 & 77.7 & 82.8 & \textbf{88.8} & 69.0 & \textbf{72.0} &  76.4 \\
			\textbf{ALDA} & & \textbf{93.8} & 74.1 & 82.4 & \textbf{69.4} & \textbf{90.6} & 87.2 & 89.0 & 67.6 & 93.4 & 76.1 & 87.7 & 22.2 & \textbf{77.8} \\
			\midrule
			Sourceonly & \multirow{3}{*}{ResNet50} & 74.6 & 26.8 & 56.0 & 53.5 & 58.0 & 26.2 & 76.5 & 17.6 & 81.7 & 34.8 & 80.3 & 27.2 & 51.1 \\
			CDAN+E~\cite{CADA} &  & - & - & - & - & - & - & - & - & - & - & - & - & 70.0 \\
			\textbf{ALDA} & & 87.0 & 61.3 & 78.7 & 67.9 & 83.7 & \textbf{89.4} & \textbf{89.5} & 71.0 & \textbf{95.4} & 71.9 & \textbf{89.6} & 33.1 & \textbf{76.5} \\
			\bottomrule
	\end{tabular}}
	\caption{Accuracy (\%)  of different unsupervised domain adaptation methods on VisDA-2017.}
	\label{tab_004}
\end{table*}

\begin{table}[t]
	\centering
	\resizebox{0.49\textwidth}{24.6mm}{    
		\begin{tabular}{ccccc}  
			\toprule
			Method & U $\to$ M & M $\to$ U &  S $\to$ M & Avg    \\
			\midrule
			Sourceonly &  77.5$\pm$0.8 & 82.0$\pm$1.2  & 66.5$\pm$1.9 & 75.3    \\
			DANN~\cite{DANN}  &  74.0 & 91.1  & 73.9 & 79.7 \\
			ADDA~\cite{ADDA}  &  90.1 & 89.4  & 76.0 & 85.2 \\
			CDAN+E~\cite{CADA}  &  98.0 & 95.6  & 89.2 & 94.3 \\
			MT+CT &  92.3$\pm$8.6 & 88.1$\pm$0.34  & 93.3$\pm$5.8 & 91.2 \\
			\multicolumn{2}{l}{\cite{self-ensembling}} & & & \\
			MCD~\cite{MCD} &  94.1$\pm$0.3 & 96.5$\pm$0.3  & 96.2$\pm$0.4 & 95.6 \\
			MCS~\cite{MCS} & 98.2 & \textbf{97.8} & 91.7 & 95.9 \\
			\midrule
			\textbf{ALDA} ($\delta=0.9$) &  98.1$\pm$0.2 & 94.8$\pm$0.1  & 95.6$\pm$0.6 & 96.2 \\
			\textbf{ALDA} ($\delta=0.8$) &  98.2$\pm$0.1 & 95.4$\pm$0.4  & 97.5$\pm$0.3 & 97.0 \\
			\textbf{ALDA} ($\delta=0.6$) &  \textbf{98.6$\pm$0.1} & 95.6$\pm$0.3  & \textbf{98.7$\pm$0.2} & \textbf{97.6} \\
			\textbf{ALDA} ($\delta=0.0$) &  98.4$\pm$0.2 & 95.0$\pm$0.1  & 97.0$\pm$0.2 & 96.8 \\
			\midrule
			Targetonly &  99.5$\pm$0.0 & 97.3$\pm$0.2  & 99.6$\pm$0.1 & 98.8 \\
			\bottomrule
	\end{tabular}}
	\caption{Accuracy (\%)  of different unsupervised domain adaptation methods on the digits datasets. We use the base model in~\cite{self-ensembling}.}
	\label{tab_001}
\end{table}

\begin{table*}[t]
	\centering
	\resizebox{0.75\textwidth}{19.0mm}{    
		\begin{tabular}{cccccccc}  
			\toprule
			Method & A $\to$ W & D $\to$ W &  W $\to$ D & A $\to$ D & D $\to$ A&  W $\to$ A & Avg    \\
			\midrule
			ResNet-50~\cite{ResNet} &  68.4$\pm$0.2 & 96.7$\pm$0.1  & 99.3$\pm$0.1  & 68.9$\pm$0.2 & 62.5$\pm$0.3 & 60.7$\pm$0.3 & 76.1    \\
			DANN~\cite{DANN}  &  82.0$\pm$0.4 & 96.9$\pm$0.2  & 99.1$\pm$0.1  & 79.7$\pm$0.4 & 68.2$\pm$0.4 & 67.4$\pm$0.5 & 82.2    \\
			\midrule
			ST  &  89.0 & \textbf{99.0}  & 100.0  & 86.3 & 67.5 & 63.0 & 84.1    \\
			DANN + ST &  91.8 & 98.4 & 100.0 & 89.1 & 68.8 & 68.7 & 86.1  \\
			ALDA w/o $\mathcal{L}_{Reg}$  &  93.8 & 98.7  & 100.0  & 91.5 & 70.4 & 67.3 & 87.0    \\
			ALDA w/o $\mathcal{L}_{T}$  &  95.0 & 97.5 & 100.0 & 94.0 & 70.8 & 69.0 & 87.7    \\
			ALDA+ST w/o $\mathcal{L}_{T}$  & 94.8  & 98.0  & 100.0  & \textbf{95.4} & 71.0 & 65.9 & 87.8    \\
			ALDA w/ $\mathcal{L}_{T}(x, \mathcal{L}_{CE})$  &  95.1 & 97.6 & 100.0 & 92.7 & 69.4 & 70.5 & 87.6    \\
			\textbf{ALDA} & \textbf{95.6$\pm$0.5} & 97.7$\pm$0.1 & \textbf{100.0$\pm$0.0} & 94.0$\pm$0.4 & \textbf{72.2$\pm$0.4} & \textbf{72.5$\pm$0.2} & \textbf{88.7} \\
			\bottomrule
	\end{tabular}}
	\caption{Ablation study on Office-31 (ResNet-50). ``ST'' denotes self-training with pseudo-labels~\cite{Self-training}.}
	\label{tab_005}
\end{table*}

\subsection{Setup}

For digits datasets, we adopt the generator and classifier networks used in~\cite{self-ensembling} and optimize the model using Adam~\cite{Adam} gradient descent with learning rate $1 \times 10^{-3}$.

For the other three datasets, we employ ResNet-50~\cite{ResNet} as the generator network. The ResNet-50 is pre-trained on ImageNet~\cite{ImageNet}. Our discriminator consists of three fully connected layers with dropout, which is the same as other works~\cite{DANN,CADA}. As we train the classifier and discriminator from scratch, we set their learning rates to be $10$ times that of the generator. 
We train the model with Stochastic Gradient Descent (SGD) optimizer with the momentum of $0.9$. We schedule the learning rate with the strategy in~\cite{DANN}: the learning rate is adjusted by $\eta_p=\frac{\eta_0}{(1+\alpha q)^{\beta}}$,  where $q$ is the training progress linearly changing from $0$ to $1$, $\eta_0=0.01$, $\alpha=10$, $\beta=0.75$. 
We implement the algorithms using \textbf{PyTorch}~\cite{PyTorch}.

There are two hyper-parameters in our method: the threshold $\delta$ of pseudo-labels and the trade-off $\lambda$. If the prediction of a target sample is below the threshold, we ignore these samples in training. We set $\delta$ to $0.6$ for digit adaptation and $0.9$ for office-31, office-home datasets and VisDA dataset. In all experiment, $\lambda$ is gradually increased from $0$ to $1$ by $\frac{2}{1+\exp (-10 \cdot q)}-1$, same as~\cite{CADA}.

\subsection{Result}

\begin{figure*}[t]
	\centering
	\begin{subfigure}[]{0.24\textwidth}
		\centering
		\includegraphics[width=0.98\textwidth]{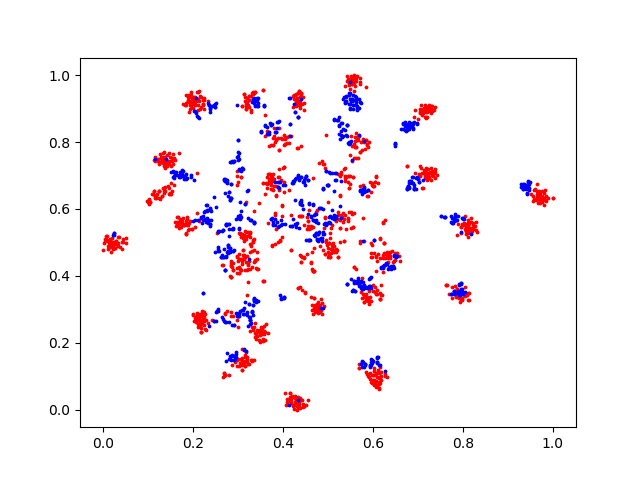}
		\caption{ResNet-50}
	\end{subfigure} 
	\begin{subfigure}{0.24\textwidth}
		\centering
		\includegraphics[width=0.98\textwidth]{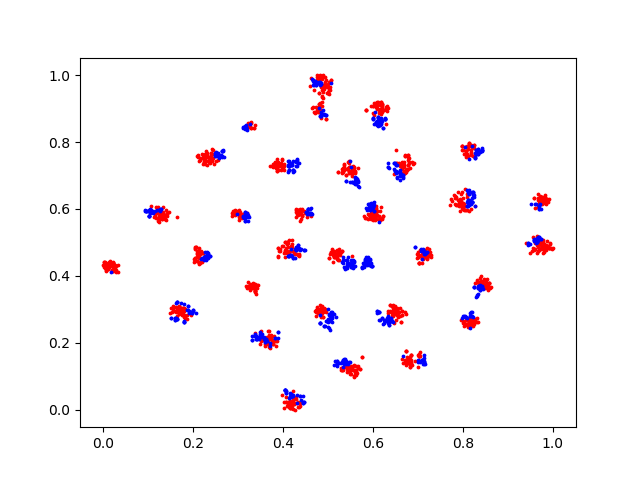}
		\caption{Self-training}
	\end{subfigure} 
	\begin{subfigure}{0.24\textwidth}
		\centering        
		\includegraphics[width=0.98\textwidth]{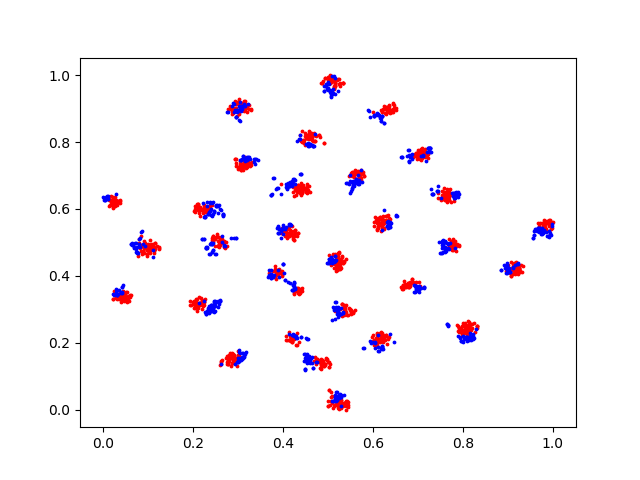}
		\caption{DANN}
	\end{subfigure} 
	\begin{subfigure}{0.24\textwidth}
		\centering
		\includegraphics[width=0.98\textwidth]{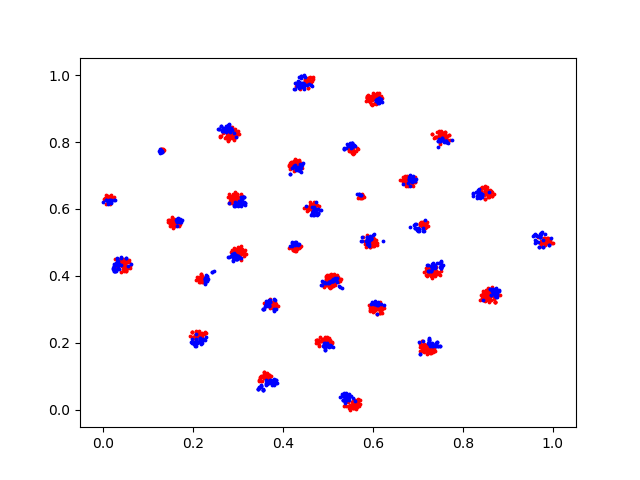}
		\caption{ALDA}
	\end{subfigure}
	\caption{T-SNE of (a) ResNet-50, (b) Self-training, (c) DANN, (d) ALDA for A $\to$ W adaptation(red: A; blue: W).}
	\label{fig_002}
\end{figure*}

\textbf{Image Results.} Table~\ref{tab_002} reports the results with ResNet-50 on Office-31. ALDA significantly outperforms state-of-the-art methods. Because ALDA combines with self-training methods to learn discriminative features, ALDA achieves better results than the domain-adversarial learning-based methods, e.g., DANN, JAN, MADA. Similar to ALDA, CDAN+E also takes the classification prediction into the discrimination and uses the entropy of prediction as an importance weight. However, 
ALDA outperforms CDAN+E on hard transfer tasks, e.g., A $\to$ W, A $\to$ D, D $\to$ A and  W $\to$ A. The outstanding results show that it is important to combine the domain-adversarial learning and self-training based methods properly.

Table~\ref{tab_003} summarizes the results with ResNet-50 on Office-home. For these more difficult adaptation datasets, ALDA still exceeds the most advanced methods. Compared to Office-31, Office-Home has more categories and has a larger appearance gap between domains. A larger number of categories indicates more components of the discriminator output $\xi$ in ALDA, which results in a stronger capacity of class-wise domain discrimination. 

Table~\ref{tab_004} shows the quantitative results with ResNet-50 and ResNet-101 on VisDA classification dataset. Even though only based on ResNet-50, our ALDA performs better than other domain adaptation methods. 

\textbf{Digits Results.} Table~\ref{tab_001} summarizes the experimental results for digits adaption comparing with state-of-the-art methods. For fair comparisons, we only resize and normalize the image and do not apply any addition data augment like ~\cite{self-ensembling}. We conduct each experiment three times and report their average results and variance. As the table shows, ALDA outperforms the most advanced distribution alignment methods, e.g., DANN, MCD, CDAN, and self-training based methods, e.g., Mean Teacher with a confident threshold (MT+CT). ALDA also reduces the performance gap between UDA and the supervised learning on the target domain by a large margin. 

In Table~\ref{tab_001}, we also investigate the effect of the threshold $\delta$ for pseudo-labels on the digits datasets. As we decrease the threshold $\delta$ from $0.9$ to $0.6$, the performances are improved. It is because the digits datasets are relatively easy to transfer and do not require high thresholds to obtain high precision pseudo-labels. The lower threshold will take more target samples into training, which promotes the training of samples with low prediction confidence. 
For the digits datasets, ALDA with $\delta=0.6$ achieves the best result.

\subsection{Analysis}

In Table~\ref{tab_005}, we perform an ablation study on Office-31 to investigate the effect of different components in ALDA. Firstly, we apply self-training~\cite{Self-training} to unsupervised domain adaptation, which is denoted as ``ST''. ``DANN+ST'' denotes that we directly combine the domain-adversarial learning and the self-training methods. However, the performance of ``DANN+ST'' is inferior to ``ALDA'', proving the importance of properly combining these two methods. To investigate the effect of the regularization term $\mathcal{L}_{Reg}$ in Eq.~\ref{equ_012}, we remove the $\mathcal{L}_{Reg}$ term in the final loss of the discriminator, denoted as ``ALDA w/o $\mathcal{L}_{Reg}$''. The results show that without $\mathcal{L}_{Reg}$, the performance of ALDA drops dramatically. This phenomenon is because the regularization term can enhance the stability of the adversarial process. 

To investigate the effect of the corrected target loss $\mathcal{L}_{T}$ in Eq.\ref{equ_015}, we remove the $\mathcal{L}_{T}$ and only keep the noise-correcting domain discrimination, denotes as ``ALDA w/o $\mathcal{L}_{T}$''. As Table~\ref{tab_005} shows, ``ALDA w/o $\mathcal{L}_{T}$'' can achieve competitive results but inferior to ``ALDA''. The phenomenon shows the superiority of our noise-correcting domain discrimination and the importance of combining domain discrimination and corrected pseudo-labels to enhance the performance. Additionally, we replace the corrected target loss $\mathcal{L}_{T}$ with uncorrected target loss, i.e., self-training with pseudo-labels, which is denoted as ``ALDA+ST w/o $\mathcal{L}_{T}$''. However, ``ALDA+ST w/o $\mathcal{L}_{T}$'' does not improve the performance, which manifests the importance of correcting pseudo-labels.

As mentioned before, the unhinged loss has been proved to be robust to the uniform part of the noise. To verify the effect of choosing the unhinged loss $\mathcal{L}_{unh}$ as basic loss function, we substitute the unhinged loss with the cross-entropy loss $\mathcal{L}_{CE}$ in the target loss $\mathcal{L}_{T}(x, \mathcal{L})$, denoted as ``ALDA w/ $\mathcal{L}_{T}(x, \mathcal{L}_{CE})$''. The results in Table~\ref{tab_005} demonstrate that the cross-entropy loss performs worse than the unhinged loss in ALDA. The unhinged loss can remove the uniform part of the noise, which facilitates the noise-correcting process.

\subsection{Visualization}

We use t-SNE~\cite{tSNE} to visualize the feature extracted by ResNet-50, Self-training, DANN and ALDA for A $\to$ W adaptation (31 classes) in Fig.~\ref{fig_002}. When using ResNet-50 only, the target feature distribution is not aligned with the source. Although self-training and DANN can align the distributions of the source and target domain, their target clusters are not fully matched with source clusters. 
For ALDA, the target clusters are closely matched with the corresponding source clusters, which demonstrates the target features extracted by ALDA are well aligned and discriminative.

\section{Conclusion}

In this paper, we propose Adversarial-Learned Loss for Domain Adaptation (ALDA) to combine the strengths of domain-adversarial learning and self-training. We first introduce the confusion matrix to represent the noise in pseudo-labels. As the confusion matrix is unknown, we employ noise-correcting domain discrimination to learn the confusion matrix. Then the target classifier is optimized with the corrected loss function. Our ALDA is theoretically and experimentally proven to be effective for unsupervised domain adaption and achieves state-of-the-art performance on four standard datasets.

\subsection*{Acknowledgments}

This work was supported in part by The National Key Research and Development Program of China (Grant Nos: 2018AAA0101400), in part by The National Nature Science Foundation of China (Grant Nos: 61936006, 61973271).

\bibliographystyle{aaai}
\bibliography{1081-aaai}

\begin{thebibliography}{}

\bibitem[\protect\citeauthoryear{Ben-David \bgroup et al\mbox.\egroup
  }{2010}]{Ben-David}
Ben-David, S.; Blitzer, J.; Crammer, K.; Kulesza, A.; Pereira, F.; and Vaughan,
  J.~W.
\newblock 2010.
\newblock A theory of learning from different domains.
\newblock {\em Machine Learning} 79(1):151--175.

\bibitem[\protect\citeauthoryear{Che \bgroup et al\mbox.\egroup
  }{2017}]{ModeRGA}
Che, T.; Li, Y.; Jacob, A.~P.; Bengio, Y.; and Li, W.
\newblock 2017.
\newblock Mode regularized generative adversarial.
\newblock In {\em ICLR}.

\bibitem[\protect\citeauthoryear{Chen, Xue, and Cai}{2019}]{Maxsquare}
Chen, M.; Xue, H.; and Cai, D.
\newblock 2019.
\newblock Domain adaptation for semantic segmentation with maximum squares
  loss.
\newblock In {\em The IEEE International Conference on Computer Vision (ICCV)}.

\bibitem[\protect\citeauthoryear{Deng \bgroup et al\mbox.\egroup
  }{2009}]{ImageNet}
Deng, J.; Dong, W.; Socher, R.; Li, L.; Li, K.; and Li, F.
\newblock 2009.
\newblock Imagenet: {A} large-scale hierarchical image database.
\newblock In {\em CVPR}.

\bibitem[\protect\citeauthoryear{French, Mackiewicz, and
  Fisher}{2018}]{self-ensembling}
French, G.; Mackiewicz, M.; and Fisher, M.
\newblock 2018.
\newblock Self-ensembling for visual domain adaptation.
\newblock In {\em ICLR}.

\bibitem[\protect\citeauthoryear{Ganin \bgroup et al\mbox.\egroup
  }{2016}]{DANN}
Ganin, Y.; Ustinova, E.; Ajakan, H.; Germain, P.; Larochelle, H.; Laviolette,
  F.; Marchand, M.; and Lempitsky, V.~S.
\newblock 2016.
\newblock Domain-adversarial training of neural networks.
\newblock {\em JMLR} 17:2096–2030.

\bibitem[\protect\citeauthoryear{Ghosh, Kumar, and Sastry}{2017}]{RobustLF}
Ghosh, A.; Kumar, H.; and Sastry, P.~S.
\newblock 2017.
\newblock Robust loss functions under label noise for deep neural networks.
\newblock In {\em AAAI}.

\bibitem[\protect\citeauthoryear{Grandvalet and
  Bengio}{2004}]{Semi-supervisedLearningbyEntropyMinimization}
Grandvalet, Y., and Bengio, Y.
\newblock 2004.
\newblock Semi-supervised learning by entropy minimization.
\newblock In {\em NIPS}.

\bibitem[\protect\citeauthoryear{He \bgroup et al\mbox.\egroup }{2016}]{ResNet}
He, K.; Zhang, X.; Ren, S.; and Sun, J.
\newblock 2016.
\newblock Deep residual learning for image recognition.
\newblock In {\em CVPR}.

\bibitem[\protect\citeauthoryear{Hong \bgroup et al\mbox.\egroup
  }{2018}]{ConditionalGANDA}
Hong, W.; Wang, Z.; Yang, M.; and Yuan, J.
\newblock 2018.
\newblock Conditional generative adversarial network for structured domain
  adaptation.
\newblock In {\em CVPR}.

\bibitem[\protect\citeauthoryear{Hull}{1994}]{USPS}
Hull, J.~J.
\newblock 1994.
\newblock A database for handwritten text recognition research.
\newblock {\em PAMI} 16:550--554.

\bibitem[\protect\citeauthoryear{Kingma and Ba}{2015}]{Adam}
Kingma, D.~P., and Ba, J.
\newblock 2015.
\newblock Adam: A method for stochastic optimization.
\newblock In {\em ICLR}.

\bibitem[\protect\citeauthoryear{LeCun}{1998}]{MINST}
LeCun, Y.
\newblock 1998.
\newblock Gradient-based learning applied to document recognition.
\newblock In {\em Proceedings of the IEEE}, volume~86,  2278–2324.

\bibitem[\protect\citeauthoryear{Lee}{2013}]{Semi-supervisedPseudo-Label}
Lee, D.-H.
\newblock 2013.
\newblock Pseudo-label : The simple and efficient semi-supervised learning
  method for deep neural networks.
\newblock In {\em ICML}.

\bibitem[\protect\citeauthoryear{Liang \bgroup et al\mbox.\egroup }{2019}]{MCS}
Liang, J.; He, R.; Sun, Z.; and Tan, T.
\newblock 2019.
\newblock Distant supervised centroid shift: A simple and efficient approach to
  visual domain adaptation.
\newblock In {\em CVPR}.

\bibitem[\protect\citeauthoryear{Liu \bgroup et al\mbox.\egroup }{2019}]{TAT}
Liu, H.; Long, M.; Wang, J.; and Jordan, M.
\newblock 2019.
\newblock Transferable adversarial training: A general approach to adapting
  deep classifiers.
\newblock In {\em ICML}.

\bibitem[\protect\citeauthoryear{Long \bgroup et al\mbox.\egroup }{2015}]{DAN}
Long, M.; Cao, Y.; Wang, J.; and Jordan, M.~I.
\newblock 2015.
\newblock Learning transferable features with deep adaptation networks.
\newblock In {\em ICML}.

\bibitem[\protect\citeauthoryear{Long \bgroup et al\mbox.\egroup
  }{2017a}]{CADA}
Long, M.; Cao, Z.; Wang, J.; and Jordan, M.~I.
\newblock 2017a.
\newblock Conditional adversarial domain adaptation.
\newblock In {\em NeurIPS}.

\bibitem[\protect\citeauthoryear{Long \bgroup et al\mbox.\egroup }{2017b}]{JAN}
Long, M.; Zhu, H.; Wang, J.; and Jordan, M.~I.
\newblock 2017b.
\newblock Deep transfer learning with joint adaptation networks.
\newblock In {\em ICML}.

\bibitem[\protect\citeauthoryear{{Mao} \bgroup et al\mbox.\egroup
  }{2017}]{LSGAN}
{Mao}, X.; {Li}, Q.; {Xie}, H.; {Lau}, R. Y.~K.; {Wang}, Z.; and {Smolley},
  S.~P.
\newblock 2017.
\newblock Least squares generative adversarial networks.
\newblock In {\em ICCV}.

\bibitem[\protect\citeauthoryear{Netzer \bgroup et al\mbox.\egroup
  }{2011}]{SVHN}
Netzer, Y.; Fillet, M.; Coates, A.; Bissacco, A.; Wu, B.; and Ng, A.~Y.
\newblock 2011.
\newblock Reading digits in natural images with unsupervised feature learning.
\newblock In {\em NIPS}.

\bibitem[\protect\citeauthoryear{Odena, Olah, and Shlens}{2016}]{ACGAN}
Odena, A.; Olah, C.; and Shlens, J.
\newblock 2016.
\newblock Conditional image synthesis with auxiliary classifier gans.
\newblock In {\em ICML}.

\bibitem[\protect\citeauthoryear{{Pan} and {Yang}}{2010}]{Survey}
{Pan}, S.~J., and {Yang}, Q.
\newblock 2010.
\newblock A survey on transfer learning.
\newblock {\em TKDE} 22(10):1345--1359.

\bibitem[\protect\citeauthoryear{Paszke \bgroup et al\mbox.\egroup
  }{2017}]{PyTorch}
Paszke, A.; Gross, S.; Chintala, S.; Chanan, G.; Yang, E.; DeVito, Z.; Lin, Z.;
  Desmaison, A.; Antiga, L.; and Lerer, A.
\newblock 2017.
\newblock Automatic differentiation in pytorch.
\newblock In {\em NIPS-W}.

\bibitem[\protect\citeauthoryear{Pei \bgroup et al\mbox.\egroup }{2018}]{MADA}
Pei, Z.; Cao, Z.; Long, M.; and Wang, J.
\newblock 2018.
\newblock Multi-adversarial domain adaptation.
\newblock In {\em AAAI}.

\bibitem[\protect\citeauthoryear{Peng \bgroup et al\mbox.\egroup
  }{2017}]{VisDA}
Peng, X.; Usman, B.; Kaushik, N.; Hoffman, J.; Wang, D.; and Saenko, K.
\newblock 2017.
\newblock Visda: The visual domain adaptation challenge.
\newblock {\em ArXiv} abs/1710.06924.

\bibitem[\protect\citeauthoryear{Saenko and Kulis}{2010}]{office}
Saenko, K., and Kulis, B.
\newblock 2010.
\newblock Adapting visual category models to new domains.
\newblock In {\em ECCV}.

\bibitem[\protect\citeauthoryear{Saito \bgroup et al\mbox.\egroup }{2018}]{MCD}
Saito, K.; Watanabe, K.; Ushiku, Y.; and Harada, T.
\newblock 2018.
\newblock Maximum classifier discrepancy for unsupervised domain adaptation.
\newblock In {\em CVPR}.

\bibitem[\protect\citeauthoryear{Saito \bgroup et al\mbox.\egroup
  }{2019}]{MinimaxEntropy}
Saito, K.; Kim, D.; Sclaroff, S.; Darrell, T.; and Saenko, K.
\newblock 2019.
\newblock Semi-supervised domain adaptation via minimax entropy.
\newblock {\em ArXiv} abs/1904.06487.

\bibitem[\protect\citeauthoryear{Sukhbaatar and Fergus}{2014}]{LearningNL}
Sukhbaatar, S., and Fergus, R.
\newblock 2014.
\newblock Learning from noisy labels with deep neural networks.
\newblock In {\em ICLR}.

\bibitem[\protect\citeauthoryear{Sun and Saenko}{2016}]{Deepcoral}
Sun, B., and Saenko, K.
\newblock 2016.
\newblock Deep coral: Correlation alignment for deep domain adaptation.
\newblock In {\em ECCV Workshops}.

\bibitem[\protect\citeauthoryear{Tarvainen and Valpola}{2017}]{Meanteacher}
Tarvainen, A., and Valpola, H.
\newblock 2017.
\newblock Mean teachers are better role models: Weight-averaged consistency
  targets improve semi-supervised deep learning results.
\newblock In {\em ICLR}.

\bibitem[\protect\citeauthoryear{{Torralba} and {Efros}}{2011}]{domainshift}
{Torralba}, A., and {Efros}, A.~A.
\newblock 2011.
\newblock Unbiased look at dataset bias.
\newblock In {\em CVPR}.

\bibitem[\protect\citeauthoryear{Tzeng \bgroup et al\mbox.\egroup }{2014}]{DDC}
Tzeng, E.; Hoffman, J.; Zhang, N.; Saenko, K.; and Darrell, T.
\newblock 2014.
\newblock Deep domain confusion: Maximizing for domain invariance.
\newblock {\em CoRR} abs/1412.3474.

\bibitem[\protect\citeauthoryear{Tzeng \bgroup et al\mbox.\egroup
  }{2017}]{ADDA}
Tzeng, E.; Hoffman, J.; Saenko, K.; and Darrell, T.
\newblock 2017.
\newblock Adversarial discriminative domain adaptation.
\newblock In {\em CVPR}.

\bibitem[\protect\citeauthoryear{van~der Maaten and Hinton}{2008}]{tSNE}
van~der Maaten, L., and Hinton, G.~E.
\newblock 2008.
\newblock Visualizing data using t-sne.
\newblock In {\em JMLR}.

\bibitem[\protect\citeauthoryear{van Rooyen, Menon, and
  Williamson}{2015}]{Unhinge}
van Rooyen, B.; Menon, A.~K.; and Williamson, R.~C.
\newblock 2015.
\newblock Learning with symmetric label noise: The importance of being
  unhinged.
\newblock In {\em NIPS}.

\bibitem[\protect\citeauthoryear{Venkateswara \bgroup et al\mbox.\egroup
  }{2017}]{officehome}
Venkateswara, H.; Eusebio, J.; Chakraborty, S.; and Panchanathan, S.
\newblock 2017.
\newblock Deep hashing network for unsupervised domain adaptation.
\newblock In {\em CVPR}.

\bibitem[\protect\citeauthoryear{Xie \bgroup et al\mbox.\egroup
  }{2018}]{LearningSemanticRepresentationsforUDA}
Xie, S.; Zheng, Z.; Chen, L.; and Chen, C.
\newblock 2018.
\newblock Learning semantic representations for unsupervised domain adaptation.
\newblock In {\em ICML}.

\bibitem[\protect\citeauthoryear{Zhang and Sabuncu}{2018}]{Nosiylabel}
Zhang, Z., and Sabuncu, M.~R.
\newblock 2018.
\newblock Generalized cross entropy loss for training deep neural networks with
  noisy labels.
\newblock In {\em NIPS}.

\bibitem[\protect\citeauthoryear{Zhang \bgroup et al\mbox.\egroup }{2018}]{CAN}
Zhang, W.; Ouyang, W.; Li, W.; and Xu, D.
\newblock 2018.
\newblock Collaborative and adversarial network for unsupervised domain
  adaptation.
\newblock In {\em CVPR}.

\bibitem[\protect\citeauthoryear{Zou \bgroup et al\mbox.\egroup
  }{2018}]{Self-training}
Zou, Y.; Yu, Z.; Kumar, B. V. K.~V.; and Wang, J.
\newblock 2018.
\newblock Unsupervised domain adaptation for semantic segmentation via
  class-balanced self-training.
\newblock In {\em ECCV}.

\end{thebibliography}

\clearpage


\section*{Supplementary material}

# Adversarial-Learned Loss for Domain Adaptation
## -Supplementary Material-

**Paper ID: 1081**

## Theoretical Proof

In the feature space $\mathcal{F}$ generated by the generator $G$, the source and target feature distributions are $\mathcal{P}_s = \{G(x_s)|x_s \in \mathcal{D}_s\}$ and $\mathcal{P}_t = \{G(x_t)|x_t \in \mathcal{D}_t\}$ respectively. If we assume that both distributions are continuous with densities $P_s$ and $P_t$, for a feature vector $f \in \mathcal{F}$, the probabilities that it belongs to source and target distributions are $P_s(f)$ and $P_t(f)$ respectively. We assume that there are an ideal classification function $y^*$ such that $y^*(f_s) = y_s, \forall f_s \in \mathcal{P}_s$. To continuously expand the function $y^*$ to the whole feature space $\mathcal{F}$, we define the function $P_{y^*}$ :

$$P_{y^*}(f) \in \mathbb{R}^K, f \in \mathcal{F}$$
$$\text{s.t. } \forall f_s \in \mathcal{P}_s, P_{y^*}(f_s)_k = 1 \text{ for } k = y_s$$
$$\text{and } P_{y^*}(f_s)_k = 0 \text{ for } k \neq y_s.$$

Similarly, we can define the function $P_{\hat{y}}$ for the classifier predictions $\hat{y}$:

$$P_{\hat{y}}(f) \in \mathbb{R}^K, f \in \mathcal{F}$$
$$\text{s.t. } \forall f \in \mathcal{P}_s \cup \mathcal{P}_t, P_{\hat{y}}(f)_k = 1 \text{ for } k = \hat{y}$$
$$\text{and } P_{\hat{y}}(f)_k = 0 \text{ for } k \neq \hat{y}.$$

To prove the Theorem 1, we need to prove a lemma firstly.

**Lemma 1.**

$$\forall x_t, \mathcal{L}_{BCE}(\mathbf{c}^{(x_t)}, \mathbf{u}^{(\hat{y}_t)}) = -2\log(1 - \xi_{\hat{y}_t}^{(x_t)}) + C \quad (1)$$
$$\forall x_s, \mathcal{L}_{BCE}(\mathbf{c}^{(x_s)}, \mathbf{y}_s) = -2\log \xi_{y_s}^{(x_t)} + C \quad (2)$$

where $C$ is a constant term that only depends on the class number $K$.

*Proof.* We first prove Eq. 1 and the proof of Eq. 2 is similarly available. As the definition, $k \in \{1,..,K\}, \mathbf{c}_k^{(x_t)} = \sum_l \eta_{kl}^{(x_t)} \hat{\mathbf{y}}_l$, where $\hat{\mathbf{y}}$ is the one-hot form of $\hat{y}$: $\hat{\mathbf{y}}_k = 1$ for $k = \hat{y}$ and $\hat{\mathbf{y}}_k = 0$ for $k \neq \hat{y}$. Therefore, $\mathbf{c}_k^{(x_t)} = \eta_{k\hat{y}}^{(x_t)}$. Because $\eta^{(x_t)}$ is *class-wise uniform* with vector $\xi^{(x_t)}$, $\mathbf{c}_k^{(x_t)} = \xi_{\hat{y}_t}^{(x_t)}$ for $k = \hat{y}$ and $\mathbf{c}_k^{(x_t)} = \frac{1-\xi_{\hat{y}_t}^{(x_t)}}{K-1}$ for $k \neq \hat{y}$.

Then
$$\mathcal{L}_{BCE}(\mathbf{c}^{(x_t)}, \mathbf{u}^{(\hat{y}_t)})$$
$$= \sum_k -\mathbf{u}_k^{(\hat{y}_t)} \log \mathbf{c}_k^{(x_t)} - (1 - \mathbf{u}_k^{(\hat{y}_t)}) \log(1 - \mathbf{c}_k^{(x_t)})$$
$$= -\log(1 - \xi_{\hat{y}_t}^{(x_t)}) - \sum_{k \neq \hat{y}_t} \frac{1}{K-1} \log(1 - \xi_k^{(x_t)}) + (K-1)$$
$$= -2\log(1 - \xi_{\hat{y}_t}^{(x_t)}) + (K-1)$$

Together with the above lemma, we can prove the theorems in the section of "Theoretical Insight".

**Theorem 1.** *When the noise-correcting domain discrimination*

$$\max_G \min_D E_{(x_s,y_s),x_t} \mathcal{L}_{Adv}(x_s, y_s, x_t) \quad (3)$$

*achieves the optimal point $D^*$ and $G^*$, the feature distributions generated by $G^*$ are aligned: $\mathcal{P}_s = \mathcal{P}_t$.*

*Proof.* We assume that there are expand functions $P_{y^*}$ and $P_{\hat{y}}$ as defined before. Then, we can expand the adversarial loss as:

$$E_{(x_s,y_s),x_t} \mathcal{L}_{Adv}(x_s, y_s, x_t)$$
$$= E_{(x_s,y_s) \in \mathcal{D}_s} \mathcal{L}_{BCE}(\mathbf{c}^{(x_s)}, \mathbf{y}_s) + E_{x_t \in \mathcal{D}_t} \mathcal{L}_{BCE}(\mathbf{c}^{(x_t)}, \mathbf{u}^{(\hat{y}_t)})$$
$$= E_{f \in \mathcal{P}_s} \mathcal{L}_{BCE}(\mathbf{c}^{(f)}, \mathbf{y}^*(f)) + E_{f \in \mathcal{P}_t} \mathcal{L}_{BCE}(\mathbf{c}^{(f)}, \mathbf{u}^{(\hat{y}(f))}),$$

where in order to transfer the expression to the feature space, we modify the notations: $\xi_k^{(f)} := \xi_k^{(x)}$, $\eta_{kl}^{(f)} := \eta_{kl}^{(x)}$, and $\mathbf{c}_k^{(f)} := \sum_l \eta_{kl}^{(f)} \hat{\mathbf{y}}(f)_l$,

For a feature vector $f \in \mathcal{P}_s \cup \mathcal{P}_t$, its contribution to the adversarial loss is:

$$\mathcal{L}(f) = P_s(f)\sum_{k=1}^{K} P_{y^*}(f)_k \mathcal{L}_{BCE}(\mathbf{c}^{(f)}, \mathbf{k}) + P_t(f)\sum_{k=1}^{K} P_{\hat{y}}(f)_k \mathcal{L}_{BCE}(\mathbf{c}^{(f)}, \mathbf{u}^{(k)})$$

$$= -2P_s(f)\sum_{k=1}^{K} P_{y^*}(f)_k \log(\xi_k^{(f)}) - 2P_t(f)\sum_{k=1}^{K} P_{\hat{y}}(f)_k \log(1 - \xi_k^{(f)}) + C$$

The above equation is derived from Lemma 1.

We take the extremum of the noise vector $\xi$ for the above formula.

$$\forall k \in \{1,..,K\}, \frac{\partial \mathcal{L}(f)}{\partial \xi_k^{(f)}}$$

$$= -2P_s(f)P_{y^*}(f)_k \frac{1}{\xi_k^{(f)}} - 2P_t(f)P_{\hat{y}}(f)_k \frac{1}{1-\xi_k^{(f)}} = 0$$

$$\implies P_s(f)P_{y^*}(f)_k(1-\xi_k^{(f)}) = P_t(f)P_{\hat{y}}(f)_k \xi_k^{(f)}$$

$$\implies \xi_k^{(f)} = \frac{P_s(f)P_{y^*}(f)_k}{P_s(f)P_{y^*}(f)_k + P_t(f)P_{\hat{y}}(f)_k}$$

The above equation is optimal point of the discriminator $D^*$. When $D^*$ can not distinguish the source and target features, we can get the optimal generator $G^*$. We take the above $\xi^{(f)}$ back into the formula of $\mathcal{L}(f)$ and compute the extremum of it. Because the induction process is too complicate to write down, we directly show the maximum point:

$$P_s(f) = P_t(f), \forall f \in \mathcal{P}_s \cup \mathcal{P}_t$$

Therefore, when the optimal $D^*$ and $G^*$ are achieved, $\mathcal{P}_s = \mathcal{P}_t$.

**Theorem 2.** *When the optimal point $D^*$ and $G^*$ are achieved in Theorem 1, if there is a optimal labeling function $y^*(f_s) = y_s, \forall f_s \in \mathcal{P}_s$ in the feature space $\mathcal{F}$, then $\forall x_t \in \mathcal{P}_t$ and $f_t = G^*(x_t)$, we have:*

$$\mathbf{c}^{(x_t)} = \begin{cases} \mathbf{h}^{y^*(f_t)} & \hat{y}_t = y^*(f_t) \\ \mathbf{u}^{(\hat{y}_t)} & otherwise \end{cases},$$

*Proof.* The optimal $D^*$ and $G^*$ are presented in the proof of Theorem 1.

$\forall x_t \in \mathcal{P}_t$, we have the discriminator outputs as:

$$\forall k \in \{1,..,K\}, \xi_k^{(x_t)}$$
$$= \frac{P_s(G^*(x_t))P_{y^*}(G^*(x_t))_k}{P_s(G^*(x_t))P_{y^*}(G^*(x_t))_k + P_t(G^*(x_t))P_{\hat{y}}(G^*(x_t))_k}$$
$$= \frac{P_{y^*}(G^*(x_t))_k}{P_{y^*}(G^*(x_t))_k + P_{\hat{y}}(G^*(x_t))_k}$$

In the case of $\hat{y}_t = y^*(x_t)$: $\xi_k^{(x_t)} = \frac{1}{2}, \forall k \in \{1,..,K\}$. Therefore, $\mathbf{c}_k^{(x_t)} = \frac{1}{2}$ for $k = \hat{y}_t$ and $\mathbf{c}_k^{(x_t)} = \frac{1}{2K-2}$ for $k \neq \hat{y}_t$. That is $\mathbf{c}^{(x_t)} = \mathbf{h}^{y^*(f_t)}$

In the case of $\hat{y}_t \neq y^*(f_t)$: $\xi_k^{(x_t)} = 1$ for $k = \hat{y}_t$ or $y^*(f_t)$, and $\xi_k^{(x_t)} = 0$ otherwise. Therefore, $\mathbf{c}_k^{(x_t)} = 0$ for $k = \hat{y}_t$ and $\mathbf{c}_k^{(x_t)} = \frac{1}{K}$ for $k \neq \hat{y}_t$. That is $\mathbf{c}^{(x_t)} = \mathbf{u}^{(\hat{y}_t)}$.

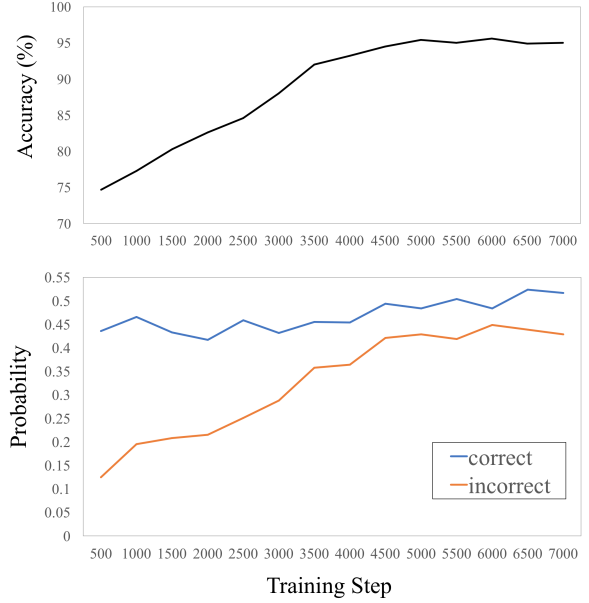

Figure 1: The upper figure: the training accuracy during training ALDA on $A \to W$. The bottom figure: the records of the weight of pseudo-labels: $\mathbf{c}_{\hat{y}_t}^{(x_t)}$ for correct samples and incorrect samples respectively.

## Experimental Analysis

To show the mechanism of the corrected target loss: $\mathcal{L}_T(x_t, \mathcal{L}) = \sum_k \mathbf{c}_k^{(x_t)} \mathcal{L}(p_t, k)$, we conduct the following experiment on the $A \to W$ adaptation with ResNet-50. During the training process of ALDA, we record the weight of pseudo-labels: $\mathbf{c}_{\hat{y}_t}^{(x_t)}$, which estimates the probability $p(y_t = \hat{y}_t | \hat{y}_t, x_t)$, for correct samples ($y_t = \hat{y}_t$) and incorrect samples ($y_t \neq \hat{y}_t$) respectively. Theoretically, Theorem 2 shows that for correct samples, we should assign large weight $\mathbf{c}_{\hat{y}_t}^{(x_t)} = \frac{1}{2}$, while for incorrect samples, we should assign small weight $\mathbf{c}_{\hat{y}_t}^{(x_t)} = 0$.

As presented in Fig.1, at the beginning of training, the weight of pseudo-labels: $\mathbf{c}_{\hat{y}_t}^{(x_t)}$ is high for correct samples ($\mathbf{c}_{\hat{y}_t}^{(x_t)} \approx 0.5$) and low for incorrect samples ($\mathbf{c}_{\hat{y}_t}^{(x_t)} \approx 0.1$). That is because, in the beginning, the correct samples are those samples that are easy to transfer and can be well matched with the source distribution by the discriminator. Consequently, the loss of pseudo-labels $\mathcal{L}(p_t, \hat{y}_t)$ will be boosted for the correct samples and surpassed for the incorrect samples, as indicated by the theorem. However, after



\end{document}